\documentclass[conference]{IEEEtran}
\IEEEoverridecommandlockouts
\usepackage{cite}
\usepackage{amsmath,amssymb,amsfonts}
\usepackage{graphicx}
\usepackage{booktabs}
\usepackage{array}
\newcolumntype{L}[1]{>{\raggedright\arraybackslash}p{#1}}
\usepackage{textcomp}
\usepackage{xcolor}
\usepackage{tikz}
\usetikzlibrary{positioning,arrows.meta,fit}
\usepackage{url}

\newcommand{\kbgym}{KBGym}

\begin{document}

\title{Training a Knowledge Base:\\Supervised Structure Learning for Agent-Curated Document Stores}

\author{\IEEEauthorblockN{Yu Pan}
\IEEEauthorblockA{\textit{University of Nebraska--Lincoln} \\
yu.pan@unl.edu}
\and
\IEEEauthorblockN{Hongfeng Yu}
\IEEEauthorblockA{\textit{University of Nebraska--Lincoln} \\
hfyu@unl.edu}}

\maketitle

\begin{abstract}
Retrieval-augmented generation treats the document store as a frozen input,
and the offline pipelines that do build structure over it build it
unsupervised --- a whole corpus indexed at uniform effort, with no signal
about which structure a question will need. We instead treat the knowledge
base as a \emph{non-parametric model} trained on (question, answer) pairs: a
\emph{curator} agent answers a supervised question against the current
store, is shown the gold answer, then edits the store. The store carries
forward, and we evaluate the curated store with a test set, on two
contamination-free benchmarks: KBGym, a fictional-universe generator we
release, and PhantomWiki. Generalization is probed with four question groups
of decreasing overlap with the training set: the trained questions
themselves, and unseen questions sharing both of their keys with training,
one key, or neither. The curated store's advantage grows with overlap ---
from parity where no key was shared, through $+0.176$ F1 where both keys
were, to $25\%$ fewer actions at $+0.294$ F1 on the trained questions, the
one cell significant on both benchmarks --- while matching HippoRAG's gains
with $1{,}913$ links against its $196{,}112$: per point of corpus covered,
$1.5\times$ the action saving and $2.1\times$ the accuracy gain. Accuracy
rises steadily with the share of the corpus the indexes cover, so training
on more questions widens coverage, and with it the generalization.
\end{abstract}


\section{Introduction}
The dominant way to give language models durable knowledge is retrieval-augmented
generation (RAG): embed a corpus once, retrieve at query time, never write back. A
rapidly growing line of work breaks this asymmetry. Agent-maintained wikis compile
sources into evolving page networks~\cite{llmwiki-rar,llmwiki-stream,karpathy-llmwiki};
agentic memory systems accumulate and link notes across sessions~\cite{amem}; and
``sleep-time'' agents reorganize memory between conversations~\cite{letta-sleep}. The
pattern has reached practice ahead of measurement: the community's own description of
the wiki pattern lists a \emph{lint} step---checking contradictions, stale claims,
orphan pages, missing cross-links---as an engineering habit with no formal evaluation
attached~\cite{karpathy-llmwiki}, and the closest published systems evaluate only
downstream answer accuracy, never the store~\cite{llmwiki-rar,llmwiki-stream}.

We study the store directly, under the oldest framing machine learning has:
\emph{train/test}. The knowledge base is the model. Supervised questions arrive one at
a time; a curator agent answers each from the current store, is graded, and then, in
an explicit consolidation phase, edits the store's structure so that the question's
\emph{class} becomes easier for a future reader. Evaluation freezes the store and
examines an independent reader with no gold access and a fixed action budget on
held-out questions.

The value produced by such training is best understood as \emph{index construction}:
a database does not change its data when an index is built, yet queries get cheaper.
Each training question's verified reasoning path is materialized as
\emph{access structure}---a multi-hop chain becomes a traversable link
path, a key that search handles badly gains an index document linking
everything under it, so future readers pay a lookup price instead of a
re-derivation price, and the one-time cost is repaid after a computable
number of questions.

\textbf{Contributions.}
\begin{enumerate}
\item \textbf{Agent-trained knowledge bases, and why supervision is the
efficient way to spend structure.} We let an LLM agent \emph{train} a
document store under the standard machine-learning paradigm, the knowledge
base playing the role of the model: supervised (question, answer) pairs are
consumed one at a time; each iteration runs a forward pass (answer from the
current store) and a backward pass (with the gold revealed, \emph{edit the
store}---adding, deleting and linking documents, building index documents);
held-out performance is the loss. The contrast with offline construction
(GraphRAG, RAPTOR, HippoRAG) is exact in machine-learning terms: those are
\emph{unsupervised}, inferring structure from the corpus with no labels and
no notion of what will be asked, while (question, answer) pairs are our
labels and structure is optimized against verified answers. The advantage
this buys is measurable and it is one of efficiency, not of ceiling. Per
point of corpus indexed, supervised curation returns $1.5\times$ the action
saving and $2.1\times$ the accuracy of an unsupervised entity index that
covers everything, because it spends its structure where questions actually
go: $1{,}913$ links against $196{,}112$, and $94\%$ of ours point at a
document that genuinely belongs under its key. And the structure generalizes, on the endpoint where generalization is
available to it: questions the store never saw gain $+0.176$ F1 when both of
their keys were indexed during training and $+0.059$ when one was, decaying
to zero when neither was. That decay is indexed by \emph{coverage}, not by
question novelty, so it recedes as training continues. Our absolute numbers
trail the unsupervised baseline only because a hundred training questions
reach a quarter of the corpus, and coverage grows linearly in questions
consumed --- the gap is training volume, not method. Supervision also composes with deployment in a
way offline construction cannot: the loop consumes questions one at a time
and edits in place, so a store can keep learning from live traffic and
concentrate its structure on the query distribution it actually serves.
\item \textbf{A train/test protocol for knowledge bases}: two-phase
(forward/backward) training iterations, single-shot gold reveal,
frozen-store examination under action budgets, learning curves, and a
\emph{key-coverage gradient} --- a probe that varies how much of a test
question's key the training set touched, which turns ``does it generalize''
into a measurable decay curve rather than a yes or no.
\item \textbf{\kbgym}, a contamination-free environment in the PhantomWiki
style~\cite{phantomwiki}: a fictional-person universe rendered as a graph of
atomic single-sentence documents; ten diagnostic question classes with exact
answers and support sets; deterministic, integer-exact store metrics and no
LLM judge in any measurement --- grading is SQuAD normalization against
programmatic golds, and the one LLM judge in the system guards
deduplication inside the store, never a reported number.
\end{enumerate}

\section{Related Work}
\textbf{Retrieval-augmented generation and agentic retrieval.} Classic RAG
retrieves from a frozen corpus with a learned dense retriever and generates
conditioned on the results~\cite{rag-lewis,dpr,realm,fid}: the corpus is an
input, never an output. Agentic RAG moves retrieval into the reasoning loop, interleaving it with
reasoning~\cite{react,ircot}, decomposing into sub-questions, retrieving on
low confidence~\cite{selfrag}, or learning the retrieve-or-reason
policy~\cite{searchr1,agentic-rag-survey}. Our reader \emph{is} such
a loop. The distinction is which side learns: that line improves the
query-side policy against a fixed corpus; we hold the policy fixed and train
the store. The axes are orthogonal and composable.

\textbf{Building structure over a corpus.} GraphRAG~\cite{graphrag},
RAPTOR~\cite{raptor}, LightRAG~\cite{lightrag}, and
HippoRAG~\cite{hipporag,hipporag2} build entity graphs, summary hierarchies,
or PageRank-linked knowledge graphs \emph{before} any question arrives.
These are our strongest baselines and the natural contrast: their structure
is unsupervised and built blind to the query distribution, ours is carved
incrementally by the questions actually asked. On the parametric side, ROME
and MEMIT edit factual associations inside model weights~\cite{rome,memit};
we pursue the non-parametric complement --- facts live in an external store
that can be inspected, reorganized, and audited, with edits that transfer
across models. STaR-style bootstrapping~\cite{star} keeps only verified
reasoning for further training; our forward/backward iteration inherits that
predict-then-learn shape, with the store rather than the weights as the
learned object.

\textbf{Agent-maintained stores and agent memory.} Recent systems have an agent
write and update wiki pages during question answering, or maintain a
time-evolving wiki under a document
stream~\cite{llmwiki-rar,llmwiki-stream,karpathy-llmwiki};
STORM~\cite{storm} generates Wikipedia-style articles by multi-perspective
research --- one-shot authorship rather than continual maintenance. A parallel line accumulates experience
across episodes --- external memory paging~\cite{memgpt}, linked long-term
note stores~\cite{mem0,amem}, reflection over an event
stream~\cite{generative-agents}, and distilled feedback or reusable
workflows~\cite{reflexion,awm,ace}. In these the
artifact is a prompt or a private memory serving one agent; ours is a
shared, inspectable document store whose curation quality is itself the
measured outcome.

\textbf{Measuring memory without contamination.} The published curation
systems report downstream task scores only: the store itself is never
measured, there is no train/test split over questions, and the evaluation
corpora are parametrically contaminated. LongMemEval and successors probe
assistant memory with QA over past sessions~\cite{longmemeval};
LoCoMo evaluates very-long-term conversational memory~\cite{locomo};
MemDelta documents hidden confounds and argues for controlled
baselines~\cite{memdelta}, supporting our no-store baseline discipline.
Multi-hop QA datasets --- HotpotQA~\cite{hotpotqa}, MuSiQue~\cite{musique},
2WikiMultiHopQA~\cite{twowiki} --- supply real-text questions with annotated
support, but their public corpora are contaminated for current models.
PhantomWiki~\cite{phantomwiki} generates contamination-free fictional
universes with programmatically exact answers; our generator follows its
recipe, and SQuAD-style normalization~\cite{squad} provides the grading.
Our protocol adds the missing measurement layer.

\definecolor{okblue}{HTML}{0072B2}
\definecolor{okorange}{HTML}{E69F00}
\definecolor{okverm}{HTML}{D55E00}
\definecolor{okgreen}{HTML}{009E73}
\definecolor{okpurple}{HTML}{9A5EA8}
\definecolor{okteal}{HTML}{56B4E9}

\begin{figure*}[t]
\centering
\begin{tikzpicture}[
  font=\small,
  node/.style={circle, draw=okblue!80, fill=okblue!25, inner sep=1.6pt},
  nav/.style={circle, draw=okverm, fill=okverm!85, inner sep=2.2pt},
  stage/.style={draw, rounded corners=2pt, align=center, inner sep=5pt},
  lbl/.style={font=\footnotesize\itshape, text=black!70},
  arr/.style={-{Stealth[length=2.2mm]}, thick, draw=black!65},
  >=Stealth]

\node[stage, draw=okblue!70, fill=okblue!5, minimum width=3.4cm,
      minimum height=2.5cm] (K0) {};
\node[above=1mm of K0.north, anchor=south] {\textbf{initial store} $K_0$};
\node[lbl, below=0mm of K0.south, anchor=north] {atomic documents, no links};
\foreach \i/\x/\y in {1/-1.1/0.6, 2/-0.3/0.8, 3/0.6/0.55, 4/-1.05/-0.1,
                      5/0.1/0.05, 6/1.05/0.0, 7/-0.6/-0.7, 8/0.35/-0.75, 9/1.0/-0.65}
  \node[node] (a\i) at ([shift={(\x,\y)}]K0.center) {};

\node[stage, draw=okorange!85, fill=okorange!6, right=1.5cm of K0, minimum width=4.2cm, align=left] (train) {
  \textbf{training iteration} $(q, a^{*})$\\[1pt]
  \textcolor{okblue!80!black}{\emph{Phase 1} forward ($N_1$): search/read $\to$ \texttt{answer}}\\
  \hspace{2mm}$\hookrightarrow$ result reveals $a^{*}$, F1\\[1pt]
  \textcolor{okverm!90!black}{\emph{Phase 2} backward ($N_2$): add/edit/delete/}\\
  \textcolor{okverm!90!black}{\hspace{2mm}link/unlink $\to$ \texttt{done}}
};
\node[lbl, below=0mm of train.south, anchor=north]
  {$\times$ 100 questions $\times$ 2 epochs};

\node[stage, draw=okgreen!80, fill=okgreen!5, right=1.5cm of train, minimum width=3.4cm, minimum height=2.5cm] (KT) {};
\node[above=1mm of KT.north, anchor=south] {\textbf{trained store} $K_T$};
\node[lbl, below=0mm of KT.south, anchor=north] {links + navigation documents};
\foreach \i/\x/\y in {1/-1.1/0.6, 2/-0.3/0.8, 3/0.6/0.55, 4/-1.05/-0.1,
                      5/0.1/0.05, 6/1.05/0.0, 7/-0.6/-0.7, 8/0.35/-0.75, 9/1.0/-0.65}
  \node[node] (b\i) at ([shift={(\x,\y)}]KT.center) {};
\node[nav] (hub) at ([shift={(-0.2,-0.35)}]KT.center) {};
\draw[okgreen!70!black, thick] (hub)--(b4) (hub)--(b5) (hub)--(b7) (hub)--(b8)
                (b2)--(b5) (b3)--(b6) (b6)--(b9);

\draw[arr] (K0) -- node[above, lbl]{} (train);
\draw[arr] (train) -- (KT);

\node[stage, draw=okpurple!80, fill=okpurple!6, below=0.9cm of KT, minimum width=4.6cm, align=center] (exam)
  {\textbf{exam} (store frozen)\\ fresh reader: search/read/\texttt{answer},
   budget $M$\\ held-out questions, gold never shown};
\draw[arr] (KT) -- (exam);
\node[stage, draw=okpurple!80, fill=okpurple!6, left=1.2cm of exam, align=center] (out)
  {F1 (non-inferior?)\\ steps per question ($\rho<1$?)};
\draw[arr] (exam) -- (out);
\end{tikzpicture}

\caption{The train/test protocol. A curator consumes supervised QA pairs,
answering each from the current store (gold revealed only after its single answer)
and consolidating verified reasoning into structure. Evaluation freezes the store and
examines an independent budgeted reader on held-out questions; the hypothesis
(Eq.~\ref{eq:hypothesis}) is non-inferior accuracy at strictly lower per-question
cost.}
\label{fig:protocol}
\end{figure*}
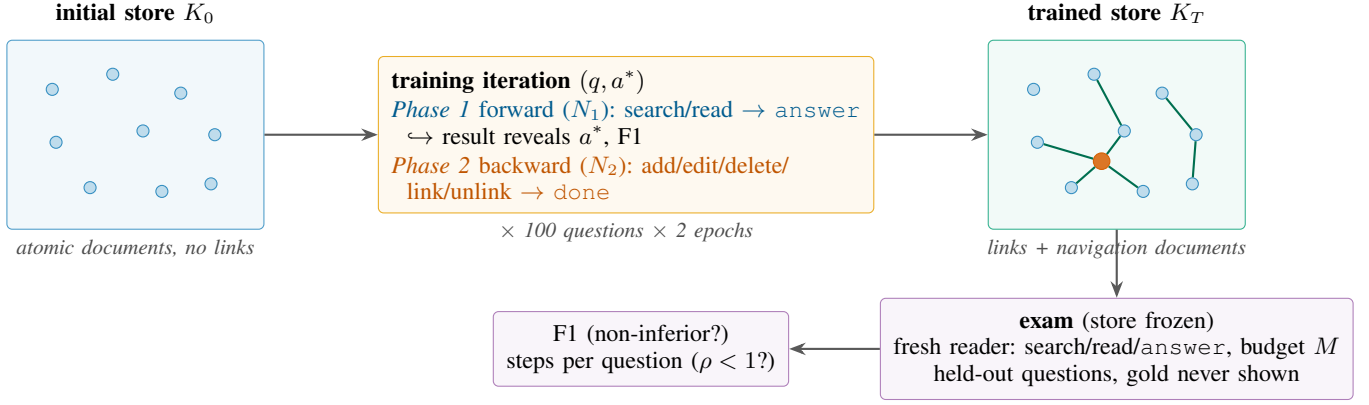

\section{Problem Formulation}
A knowledge base $K$ is a directed graph of \emph{documents}, each one
self-contained sentence plus its outgoing edges, so that what the store
learns is visible in its structure rather than hidden inside prose. Two
fields are all any agent sees or writes: \texttt{text}, the one sentence
\texttt{search} matches against, and \texttt{links}, the edges
\texttt{read} follows --- \texttt{read} returns a document together with the
full text of everything it links, which is what lets a well-connected
document replace a sequence of uncertain searches. An \emph{index document}
is not a distinct type but a document whose text names a key and whose worth
lies in its links; nothing marks it as one, which is why an index carrying no
links is indistinguishable from a sentence nobody needs. Three further
fields are kept by the environment and are invisible to the agent, so that
measurement does not depend on its cooperation: \texttt{origin} (which
original fact a document represents, empty when authored), \texttt{flag}
(untouched / authored / edited), and \texttt{absorbed} (origins folded in by
a merge). Together they make the ledger exact.

The \emph{reader} is a fixed LLM policy with \texttt{search}, \texttt{read}
and \texttt{answer}, a fixed action budget $M$, and a memory holding only the
current question and the last $k$ action--result pairs. Its behaviour on a
question distribution $Q$ defines store quality on two axes,
$\mathrm{acc}(K) = \mathbb{E}_{q}[\mathrm{F1}(\mathrm{reader}(K,q))]$ and
$\mathrm{cost}(K) = \mathbb{E}_{q}[\mathrm{steps}(\mathrm{reader}(K,q))]$.
\emph{Training} is any procedure consuming supervised pairs $(q,a^{*})$ and
editing $K$, producing $K_T$ from $K_0$.

What such a procedure should learn is \emph{access structure}, and the
distinction matters because the action set admits a shortcut that resembles
learning. A store can gain \emph{content} --- a document stating a fact,
including one the store already implies, such as an answer just verified ---
or \emph{access structure}: index documents whose value is the links they
carry, and edges a chain must step across. Only the second generalizes. A
document recording a verified answer serves the instance that produced it,
and that instance will not recur; an index over a key serves every question
whose search names that key. Our hypothesis is therefore a claim about
access structure, and we measure navigability alongside accuracy and cost:
\begin{equation}
\mathrm{acc}(K_T) \;\ge\; \mathrm{acc}(K_0) - \delta
\quad\text{and}\quad
\rho \;=\; \frac{\mathrm{cost}(K_T)}{\mathrm{cost}(K_0)} \;<\; 1,
\label{eq:hypothesis}
\end{equation}
with margin $\delta = 0.03$. Training amortizes: for $C_{\mathrm{train}}$
tokens spent and a per-question saving of $c_0 - c_T$, the break-even point
\begin{equation}
N^{*} \;=\; C_{\mathrm{train}} / (c_0 - c_T)
\label{eq:breakeven}
\end{equation}
is the query volume after which curation has paid for itself.

The correspondence to supervised learning is exact: the store configuration
is the parameter vector, one iteration is one optimization step, the
curator is the optimizer, and held-out reader performance is the
loss. Because training forward passes and test readers use identical tools,
budgets and the single-shot answer rule, the generalization gap is well
defined --- and Sec.~\ref{sec:res-grad} refines it from a train/test
dichotomy into a gradient over how much of a question the training set
touched.

\section{The \kbgym{} Environment}\label{sec:env}
\subsection{Universe, Store and Questions}
A generator in the PhantomWiki style samples a fictional population (500
people): family trees, spouses, friendships, per-person attributes (job,
hobby, city), distinct birthdates, and \emph{name distractors} sharing first
names with questioned subjects. Facts render through fixed templates into
5{,}864 atomic single-sentence documents (``Alice Johnson's job is
arborist.''), and the initial store has \emph{zero links}: a bag of facts.
Ten question categories over 26 templates. QC1--QC3 are 1--3-hop
named-entity chains, natively easy for document-level retrieval, and serve
as the non-inferiority floor. The rest target relations \emph{between}
documents: aggregation counts (QC4, QC8), abstention (QC5), multi-constraint
joins with no name anchor (QC6), set intersection (QC7), superlatives over
birthdates (QC9), reverse lookup with a uniqueness guarantee (QC10). Golds
come from graph traversal, every question carries its support set, and
grading is SQuAD-normalized F1. Splits are instance-disjoint:
\texttt{train} 150, \texttt{test\_in} 100 (unseen instances of trained
templates), \texttt{test\_out} 50 (one reserved template per category),
\texttt{eval} 30 (drives the per-epoch curve so the test splits are touched
once).

\subsection{Actions}\label{sec:actions}
Table~\ref{tab:actions} is the complete action set. Availability is enforced
by the tool schema rather than at runtime: the reader is never shown an
editing tool, so it cannot decline to use one, and the curator is never
shown \texttt{answer}. The reader's three actions are byte-identical in
training phase~1 and at exam time, which is what makes the frozen store the
only thing that differs between the two.

Two prices are set deliberately. \texttt{search} returns five documents per
page and a page costs an action, so enumerating a set is possible but
expensive --- the twenty-nine residents of a city cost six actions, while one
\texttt{read} of a complete index costs one. Closing that gap is exactly
what a trained store is for, and returning sixty results at once would erase
the thing being measured. Conversely \texttt{link\_many} attaches up to forty
targets for a single action, so building an index is cheap once its
members are found: the curator's budget goes on finding, not on attaching.
Batches above forty targets are rejected --- the largest genuine key in this
universe has thirty-six members, so a larger batch is a search result rather
than a set, and truncating it silently would leave the curator believing it
had built something it had not.

Provenance is tracked throughout: initial documents carry an origin, edited
ones are flagged, authored ones have no origin. Coverage and duplication are
therefore integer-exact and any authored content is attributable.

\begin{table}[!t]
\caption{The action set. \emph{R} = available to the reader (exam and
training phase~1); \emph{C} = available to the curator (training phase~2).
No action is available to both roles in a way that lets the reader modify
the store.}
\label{tab:actions}
\centering
\scriptsize
\renewcommand{\arraystretch}{1.15}
\setlength{\tabcolsep}{4pt}
\begin{tabular*}{\columnwidth}{@{\extracolsep{\fill}}lL{4.3cm}cc@{}}
\toprule
Action & Effect & R & C \\
\midrule
\texttt{search(q, page)} & five most similar documents; one page per action & \checkmark & \checkmark \\
\texttt{read(id)} & the document and the full text of everything it links to, one level & \checkmark & \checkmark \\
\texttt{answer(text)} & submit and end the pass & \checkmark & \\
\midrule
\texttt{add(text)} & new document; a near-duplicate is merged instead & & \checkmark \\
\texttt{edit(id, text)} & replace a document's text & & \checkmark \\
\texttt{delete(id)} & remove a document & & \checkmark \\
\texttt{link(a, b)} & one directed edge & & \checkmark \\
\texttt{link\_many(a, T)} & up to forty edges from \texttt{a} for one action & & \checkmark \\
\texttt{unlink(a, b)} & remove an edge & & \checkmark \\
\texttt{done()} & end the curation pass & & \checkmark \\
\bottomrule
\end{tabular*}
\end{table}

\subsection{Design Rationale}\label{sec:design}
Four decisions carry the methodology. \emph{(i) No LLM judge in any
measurement}: golds are exact and store metrics are integer counts over
provenance, so every number here is reproducible to the digit; the one judge
in the system vetoes near-duplicate writes and scores nothing.
\emph{(ii) Realistic retrieval}: \texttt{search} returns matched document
text, as production vector stores do --- an early variant returning only
titles collapsed the reader to F1 $0.0$ and would have credited the trained
store for repairing a crippled interface. \emph{(iii) Atomic documents}: the
store is its own chunking, so organization is expressible \emph{only}
through links and index documents, and any efficiency gain is attributable
to structure. \emph{(iv) Contamination control}: gpt-5-mini answers
HotpotQA-style questions at F1 $\approx 1.0$ with \emph{no retrieval at
all}, so on public corpora a reader's score conflates store with weights. A
fictional universe puts the no-store score at $\approx 0$, and every point
of reader performance is earned through the store.

\section{Training Protocol}
Fig.~\ref{fig:protocol} gives the shape of one iteration and of the exam
that follows.

Phase 1 uses exactly the reader's toolset and budget, so train-forward accuracy is
directly comparable to test accuracy. The single-shot \texttt{answer} reveals the gold
in its result---standard supervised learning: predict, then see the label, then
update. Leftover Phase-1 budget is forfeited (no smuggling forward steps into
consolidation); a budget-exhausted forward scores $0$ but still receives the gold, so
failed questions---where repair matters most---get targeted consolidation.
Every intermediate store state is reconstructible by replaying the edit trace
(validated byte-exact against epoch snapshots), giving per-iteration store
trajectories for analysis at no storage cost.

What the backward pass is told is deliberately uniform: no oracle
localization, and no branching on how the forward pass failed. The agent
receives its own trajectory, the gold answer, its F1, and one of three
coarse outcomes (budget exhausted, answered wrongly, answered correctly),
followed by the same instruction in every case --- name the keys this
question mentioned, the people, places, jobs, hobbies and relations it
named, and make sure each has a complete index document: build what is
missing, extend what is partial, change nothing if they are already
complete.

The curation skill supplies the standing constraints rather than a
procedure. An index \emph{points}; its members belong in its links, never
recited in its sentence, so the index survives the facts changing under it.
One key at a time: a question naming four keys and a budget that closes one
should close one, because an index of twenty-nine members is worth more than
four of three, and an unreached key is picked up by the next question naming
it. Precision before completeness: search returns what is similar, not what
belongs, so linking a whole result set destroys the certainty an index
exists to provide. Completeness is then a promise --- a reader that finds an
index stops searching, so a partial index does not merely underperform, it
makes the reader confidently count nine where there are fourteen. And an
index under which the store holds nothing is deleted rather than left
standing, since it would cost a retrieval slot and return nothing.
Parsimony governs all of it: duplicated documents compete in search and bury
each other. That last constraint encodes the lesson of preliminary runs on
an earlier environment variant, in which an agent trained \emph{without} it
quintupled an already-organized store and regressed the reader by 17 F1
points while every individual edit looked constructive.

Credit assignment needs no oracle: a wrong answer plus the revealed gold
lets the agent re-search \emph{with the answer in hand}, and what to repair
follows from the difference between where it searched and where the answer
turned out to live.

\section{Experimental Design}
\subsection{Benchmarks}
Two arms share one protocol and differ in who authored the questions
(Table~\ref{tab:arms}).

\begin{table}[t]
\caption{Benchmark arms. Both use the same splits scheme, reader, action set and budgets; the arms differ in who wrote the questions.}
\label{tab:arms}
\centering
\scriptsize
\renewcommand{\arraystretch}{1.2}
\begin{tabular*}{\columnwidth}{@{\extracolsep{\fill}}L{1.7cm}L{2.4cm}L{2.4cm}@{}}
\toprule
 & \kbgym{} (main) & PhantomWiki (official) \\
\midrule
documents & 5{,}864 sentences & 3{,}403 sentences from 405 articles \\
source & our generator & their generator \\
questions & 330 (150/100/50/30) & 330, same scheme \\
question author & us & PhantomWiki \\
supports & exact & unavailable \\
contamination & none & none \\
\bottomrule
\end{tabular*}
\end{table}

\textbf{Official PhantomWiki (external validity).} A generator we do not
control (phantom-wiki 1.0.3): 16 family trees, 405 person articles, 480
generated QA over 8 templates in four composition shapes, hop depth 1--5;
two reserved templates form \texttt{test\_out}. Its Prolog-derived relations
carry no per-fact supports, so repair diagnostics are unavailable there,
which the protocol itself does not need. A real-text arm is the obvious
third benchmark and we did not run it: public corpora are parametrically
contaminated, so every score would need a no-store baseline that is a study
in itself.

\subsection{Baselines}
All baselines are \emph{unsupervised}: structure is induced from the corpus
alone, with no access to questions or answers --- the defining contrast with
our curator, for which gold answers are labels. They differ from us and from
each other \emph{only} in how the store is prepared; reader, test sets and
budgets are identical. \textbf{B1, flat store}: the untrained store
(= epoch~0), no structure, no build cost. \textbf{B2,
GraphRAG-style}~\cite{graphrag}: lexically extracted entities per document,
co-occurrence communities, one LLM-written summary document per community
linked to its members (488k build tokens). \textbf{B3,
HippoRAG-2-style}~\cite{hipporag2}: (subject, relation, object) triples per
document, entity-sharing links, per-entity hub documents (lexical, so
$\approx$0 build cost). Both land their structure as documents and links
\emph{in our store format}, so the common reader can traverse it: we test
their \emph{structure} under a fixed reader rather than their retrieval
algorithms, a documented adaptation --- B3 forgoes Personalized PageRank.

\subsection{Implementation}
\textbf{Storage and retrieval.} The store is a dictionary from document id
to a record of \texttt{text} and outgoing ids; there is no database, links
are ids rather than copies, and a delete cascades so no edge dangles. A
separate in-memory vector index (chroma, default ONNX MiniLM embeddings)
serves \texttt{search} and holds nothing else: adding or editing a document
marks it dirty, and the environment re-embeds dirty documents in one batch at
the end of an iteration, so the index is strictly derived state.
\textbf{Reproducibility.} Every action and its result is appended to a
trace, so any intermediate store is rebuildable exactly by replay --- without
an API call, including merges, whose verdicts are recorded rather than
recomputed. This is not only convenient: when an API outage killed a run 100
iterations in, the store was recovered from its trace in under two minutes.

\textbf{Models and budgets.} Curator and reader are gpt-5-mini (temperature
0.3, low reasoning effort, 12k completion tokens per turn) with tool-forced
decisions and static schemas; the \texttt{gpt-5-mini} alias resolved to
snapshot \texttt{gpt-5-mini-2025-08-07}, which we record because the alias
will move. Budgets $N_1{=}M{=}15$, $N_2{=}30$ --- the backward pass is longer
because it must both locate a set and attach it --- and FIFO memory $K{=}30$.
The main run averaged 94k tokens and 123\,s per iteration (200 iterations,
18.8M tokens, 6.8\,h, $\approx$\$13), single seed.

\textbf{The deduplication guard.} On add or edit the store embeds the
candidate, takes the three nearest documents, and consults an LLM judge when
cosine similarity exceeds $0.90$. The judge gets 500 completion tokens at
minimal reasoning effort, which matters more than it sounds: a reasoning
model spends its budget thinking before it emits anything, and at the four
tokens an earlier version allowed it returned an empty string on every call,
read by the caller as ``not a duplicate'' --- the guard was wired up and
inert. Measured directly, 4 and 64 tokens both yield empty output and 500
answers in about ten, so the merge counts reported here come from a judge
that answered.

\subsection{Experiment Suite}\label{sec:suite}
Four experiments follow, each stated with the prediction it was designed to
falsify. E1 reads the training dynamics; E2, the headline, is the
key-coverage gradient, whose prediction is that $\rho$ rises monotonically
from the trained group towards $1$ and that the rate at which it rises
\emph{is} the transfer radius; E3 decomposes retention by question class,
predicting that gains concentrate in the relation-level classes retrieval
cannot solve natively; E4 audits the store itself. Break-even $N^{*}$
(Eq.~\ref{eq:breakeven}) closes the analysis.

\section{Experiment Results}\label{sec:prelim}
All results are from the final environment and the main run.

\subsection{E1: Training Dynamics}\label{sec:res-e1}
The 500-person configuration was chosen by probing untrained stores: forward
chains are natively easy at any scale (F1 $1.0$), while the relation-level
classes leave headroom that grows with the universe (untrained F1 $0.83$ at
120 people against $0.67$ at 500, mean steps $6.7$ against $9.2$), so the
cost axis has roughly $3\times$ room above its $\sim$3-step floor.
The main run trains on 100 questions for two epochs (200 iterations, 18.8M
tokens, 6.8\,h, $\approx\$13$). Two signals establish that the store
accumulates before the frozen exams test what it is worth. Forward accuracy
climbs \emph{within} epoch~1, from $0.60$ over the first fifty iterations to
$0.66$ over the last fifty, before any question repeats: later training
questions already benefit from structure earlier ones left behind. And
curation slows as the store fills --- $4.9$ edits per iteration in epoch~1
against $1.6$ in epoch~2, $25.3$ of $30$ backward actions spent against
$14.7$, document growth $2.4$ per iteration against $0.45$ --- because the
agent increasingly finds what a question names already indexed, which is
what the parsimony objective asks for. The per-epoch eval curve is flat
(eval-in $0.75 \to 0.80 \to 0.80$, eval-out $1.00 \to 0.90 \to 0.90$,
$n{=}20$ and $10$); at that sample size it is not distinguishable from
noise, and the eval split is instance-disjoint from training, which places
it at the far end of the coverage gradient measured next. We report it
because it was pre-registered, not because it carries weight.

\subsection{E2: The Key-Coverage Gradient}\label{sec:res-grad}
\begin{figure*}[!t]
\centering
\includegraphics[width=0.82\textwidth]{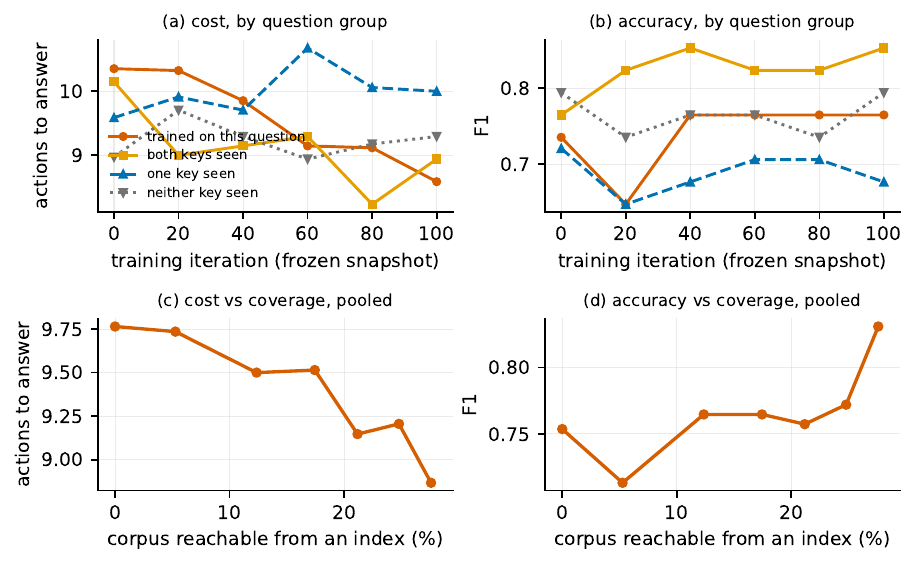}
\caption{Six frozen snapshots from epoch~1, read two ways. The reader,
action set, budget and questions are identical at every point, so the only
thing varying is the store. \emph{(a, b)} Cost and accuracy per question
group against training iteration: the two key-covered groups improve while
the two uncovered ones do not --- gains track key coverage, not question
identity. \emph{(c, d)} The same runs pooled, against the share of the
corpus the store's indexes reach --- the quantity training actually buys.
Cost falls and accuracy rises with coverage; the last point is the epoch-2
store. Coverage reaches $24.8\%$ by the end of epoch~1; the second pass
(Sec.~E2) then converts that structure into most of the accuracy gain.}
\label{fig:learning}
\end{figure*}

How far does built structure reach? A train/test split answers that with a
single number, which conflates two very different failures: a store that
memorized its training questions and a store that generalizes but was never
given enough of the corpus to cover the test set. We separate them by
constructing four question groups that differ in how much of the question
the training set touched. \kbgym{} is generated, so the full instance pool
of every template is available, and the groups are \emph{stratified}: all
four draw from the six two-slot trained templates with identical
per-template quotas ($n{=}34$), so no group is easier by composition ---
the anchor group is every two-slot training question, and the other three
mirror its template mix with unseen questions whose keys training touched
twice, once, or not at all, all examined on the same frozen store, reader
and budget. Trained questions still name the more popular keys, so B1's
own F1 varies across groups ($0.62$--$0.79$) and the table reads
\emph{vertically}: each arm against B1 on the same questions.

Table~\ref{tab:gradient} is the result and Fig.~\ref{fig:learning} its
training-time version: on the stratified probe the decay is monotone on
both endpoints. On trained questions the reader spends $25\%$ fewer
actions ($\rho = 0.747$, CI $[0.64, 0.87]$) at $+0.294$ F1 over the flat
store; with both keys covered, $+0.176$ F1 and the step saving loses
significance ($\rho = 0.879$); at one key $+0.059$; at none, nothing.
Accuracy decays with key coverage; the step saving, significant only
where the exact question was trained, decays faster.

\textbf{Two epochs, two different gains.} Fig.~\ref{fig:learning}a--b
separates them. Across epoch~1, where each question is seen once, the
trained group's \emph{cost} moves most of the way it is going to move
($10.35 \to 8.59$ actions) while accuracy barely does ($0.735 \to 0.765$).
Across epoch~2, where the same hundred questions are curated a second
time, accuracy jumps ($0.765 \to 0.912$ on the trained group,
$0.853 \to 0.941$ one step out) while cost only edges further
($8.59 \to 8.00$). The first pass builds the navigation that makes
answers cheaper; the second pass is what makes them \emph{right} --- and
it does not work by completing indexes left half-built (mean out-degree is
flat across the epoch, $6.79$ to $6.67$, and the empty count barely moves,
$45$ to $43$) but by adding $45$ more indexes aimed at the same questions.
The gains we report therefore reflect two curation passes over a question,
not one, which is a real qualification on the headline numbers and a
direct argument for the online setting: a store serving repeat traffic
gets more passes over the keys that matter, for free.

\textbf{The transfer radius depends on which endpoint you ask about.}
Splitting each group by whether the flat store exhausted its budget
separates two effects that the pooled $\rho$ hides. Where B1 runs out of
actions, the trained store both finishes sooner and answers better, and this
persists past the exact key and decays with it: F1 rises $0.08 \to 0.77$
on the trained questions, $0.12 \to 0.88$ at two keys covered,
$0.14 \to 0.50$ at one, and only $0.12 \to 0.25$ at none. Where B1
already finishes comfortably, the trained store is slightly \emph{slower}
off the trained set ($\rho$ between $1.08$ and $1.20$), because its
indexes occupy retrieval slots on questions that did not need them. Built
structure transfers as accuracy on questions the flat store cannot finish,
and as step savings only on the questions it was trained on. Reporting one
number for ``does it generalize'' would have hidden both halves.

\textbf{The same shape on questions we did not write.} \kbgym{} is our
generator, so the gradient could be an artifact of templates chosen by the
people who designed the method. The PhantomWiki arm answers that: its
universe (3{,}403 documents from 405 articles) and every one of its
questions come from a generator we do not control, and the protocol runs
unchanged (200 iterations, 14.9M tokens). Its questions carry at most one
key, so the gradient there has three points rather than four, and they
align with \kbgym{}'s:

\begin{center}\small
\begin{tabular}{lccc}
\toprule
$\rho$ (F1) & trained on & one key & neither \\
\midrule
\kbgym{} & 0.747$^{*}$ \emph{(0.912)} & 0.977 \emph{(0.706)} & 1.016 \emph{(0.765)} \\
PhantomWiki & 0.769$^{*}$ \emph{(0.880)} & 0.914 \emph{(0.807)} & 1.029 \emph{(0.718)} \\
\bottomrule
\end{tabular}
\end{center}

The trained group is the only cell significant on both benchmarks
(CIs $[0.64,0.87]$ and $[0.68,0.86]$), and accuracy rises on both
($0.618 \to 0.912$ and $0.728 \to 0.880$). Neither benchmark shows a
significant step saving off the trained set: partial transfer lives in the
accuracy endpoint, while the step endpoint pays out only where the exact
question was trained. On PhantomWiki the
uncovered group is worse than neutral --- F1 falls $0.765 \to 0.718$ --- as
indexes for other keys occupy retrieval slots that question needed. Coverage
is not merely absent outside the trained set; it is mildly costly, which is
the sharpest argument for extending it rather than accepting a quarter of
the corpus. The full four-point gradient is unavailable on this arm because
constructing one requires enumerating a template's entire instance pool and
PhantomWiki ships the questions its generator produced rather than the pool
behind them; the arm also carries no offline-construction baselines, so the
comparison below is \kbgym{}-only.

\textbf{Coverage is the currency.}\label{sec:res-perlink}
B3 answers in fewer actions and at higher F1 than we do, from $100\%$
coverage against our $27.6\%$: it indexes every entity in the corpus, while
a hundred training questions name a quarter of it. Its $\rho$ holds
between $0.70$ and $0.81$ in all four groups --- flat, because every group
meets the same fully-indexed store --- while ours tracks what training
touched, matching B3's accuracy where both keys were covered and falling
back to the flat store where neither was. A gradient defined by what training touched is in
any case a property of our arm alone, so the comparison that treats both
fairly divides each arm's gain by the coverage that produced it.

\emph{Coverage} counts a source document as covered when at least one
authored index links to it directly, and is reported as a share of the
5{,}864 originals. Three alternative readings agree, so the number is not an
artifact of the definition: allowing a second hop through an index-to-index
edge leaves it unchanged at $27.6\%$ (the agent built only 100 such edges);
counting only links a semantic check confirms as correct gives $27.3\%$; and
at the level of questions rather than documents, $78$ of $302$ ($25.8\%$)
have their entire support set reachable from some index in one read. Both
offline arms sit at $100\%$ on every one of these readings, by construction.

Pooled over all 136 probe questions, B1 answers in $9.8$ actions at F1
$0.706$. B3, indexing the whole corpus, reaches $7.5$ actions and $0.919$;
our store, indexing $27.6\%$ of it, reaches $8.9$ and $0.831$. Per point
of corpus coverage that is $0.035$ actions saved and $+0.0045$ F1 for us
against $0.023$ and $+0.0021$ for B3 --- $1.5\times$ and $2.1\times$ more
per point covered. Both ratios concentrate where training touched the
keys (at one key the accuracy ratio falls below $1$; at none ours is
negative), so the per-point advantage is worth whatever a deployment's
overlap with its training distribution makes it worth. B2, which also
covers everything, returns \emph{negative} actions and F1 per point:
coverage alone is not the mechanism. What separates us from B3 is not the
quality of the structure but how much of the corpus carries any.

\begin{table}[!t]
\caption{What a point of corpus coverage buys. Each arm's gain over the
untrained flat store, pooled over all 136 probe questions, divided by the
share of the corpus its indexes reach. The offline arms index everything;
ours indexes the part its training questions named.}
\label{tab:percoverage}
\centering
\scriptsize
\renewcommand{\arraystretch}{1.2}
\setlength{\tabcolsep}{5pt}
\begin{tabular*}{\columnwidth}{@{\extracolsep{\fill}}lrrr@{}}
\toprule
Store & coverage & actions / pt & F1 / pt \\
\midrule
B1 flat & 0\% & --- & --- \\
B2 GraphRAG & 100\% & $-0.002$ & $-0.0003$ \\
B3 HippoRAG2 & 100\% & $0.023$ & $+0.0021$ \\
Ours (trained) & 27.6\% & $\mathbf{0.035}$ & $\mathbf{+0.0045}$ \\
\bottomrule
\end{tabular*}
\end{table}

\textbf{The store is undertrained, not saturated.} This is the central
qualification on every number in this paper, and Fig.~\ref{fig:learning}c--d
is the evidence for it. Through epoch~1 the
reachable share of the corpus grows almost linearly in questions consumed,
at $0.25$ points per question. Epoch~2 re-asks the same hundred questions
and the curve flattens immediately --- $24.8\%$ to $27.6\%$ over a hundred
further iterations. Nothing saturated; the supply of new keys ran out.
Extending the epoch-1 slope reaches full coverage at roughly $400$ distinct
training questions and $\approx 48$M tokens --- about four times the
training we ran, rather than a change of method. That projection is a lower
bound: the largest keys are hit first, so later questions cover less each,
and we state it as an extrapolation rather than a result.

\textbf{The protocol is already the online one.} Training consumes questions
one at a time and edits in place; nothing in the loop needs the question set
in advance, and the frozen-store exam is a measurement device rather than a
deployment constraint. A store curated against live traffic would therefore
accumulate coverage on exactly the keys its users ask about --- the
distribution where, by Table~\ref{tab:gradient}, coverage is worth the most.
Offline construction cannot follow a query distribution it never sees. We
did not run that experiment: separating a training phase from a frozen exam
is what makes the generalization question answerable at all, and an
always-learning store cannot be said to have been tested on anything. The
two readings are complementary, and the online one is the deployment we
think this protocol is actually for.

\begin{table}[!t]
\caption{The key-coverage gradient. Groups differ in how much of the
question the training set touched, stratified so all four share one
template mix; residual key-popularity differences remain, so compare
within a column: each arm against B1 on the same questions. $\rho$ =
actions relative to the untrained flat store on the same questions, so
lower is cheaper; $^{*}$ marks a bootstrap 95\% CI on $\rho$ excluding
$1$. $n=34$ per group,
identical reader, action set and budget $M{=}15$. HippoRAG2 is absent
because it indexes the whole corpus, so all four groups meet the same store
and the gradient cannot separate them; it is compared in
Table~\ref{tab:percoverage} instead.}
\label{tab:gradient}
\centering
\scriptsize
\renewcommand{\arraystretch}{1.15}
\setlength{\tabcolsep}{4pt}
\begin{tabular*}{\columnwidth}{@{\extracolsep{\fill}}lrrrr@{}}
\toprule
Store & trained on & both keys & one key & neither \\
\midrule
\multicolumn{5}{@{}l}{\emph{$\rho$: actions relative to B1}}\\
\quad B1 flat & 1.000 & 1.000 & 1.000 & 1.000 \\
\quad B2 GraphRAG & 0.918 & 1.154 & \textbf{0.972} & 1.047 \\
\quad Ours (trained) & \textbf{0.747}$^{*}$ & \textbf{0.879} & 0.977 & \textbf{1.016} \\
\addlinespace[2pt]
\multicolumn{5}{@{}l}{\emph{F1}}\\
\quad B1 flat & 0.618 & 0.765 & 0.647 & \textbf{0.794} \\
\quad B2 GraphRAG & 0.618 & 0.676 & \textbf{0.706} & 0.706 \\
\quad Ours (trained) & \textbf{0.912} & \textbf{0.941} & \textbf{0.706} & 0.765 \\
\bottomrule
\end{tabular*}
\end{table}

\subsection{E3: Retention by Question Class}\label{sec:res-e3}
On the same 100 questions seen a second time, forward F1 rises $0.63 \to
0.82$ and the movement is entirely in the relation-level classes: the
forward chains were saturated from the start (QC1--QC3 and QC9 at $1.00$ in
both epochs, since document-level retrieval already solves them), while
counts QC4 go $0.33 \to 1.00$, joins QC6 $0.33 \to 0.92$, reverse lookup
QC10 $0.20 \to 0.60$, intersection QC7 $0.67 \to 0.83$, deep counts QC8
$0.42 \to 0.58$ and abstention QC5 $0.00 \to 0.25$ ($n$ between 5 and 14 per
class). The gains land exactly in the classes retrieval cannot solve
natively. Cost falls too --- $7.4$ steps and 70k tokens against $8.4$ and
118k, budget exhaustion $15/100$ to $8/100$, and on the 83 questions
resolved within budget in \emph{both} epochs the lookup shortens $7.16 \to
6.37$ steps while F1 rises $0.759 \to 0.892$, so the saving is not an
artifact of more questions terminating early. These two epochs are not a
controlled store contrast, though: the store evolves \emph{during} epoch~1,
so the epoch label mixes store state with question order. The controlled
version is the \emph{trained} column of Table~\ref{tab:gradient}, which
gives a larger saving ($\rho = 0.747$) precisely because it removes that
mixing.

\begin{figure*}[!t]
\centering
\includegraphics[width=0.86\textwidth]{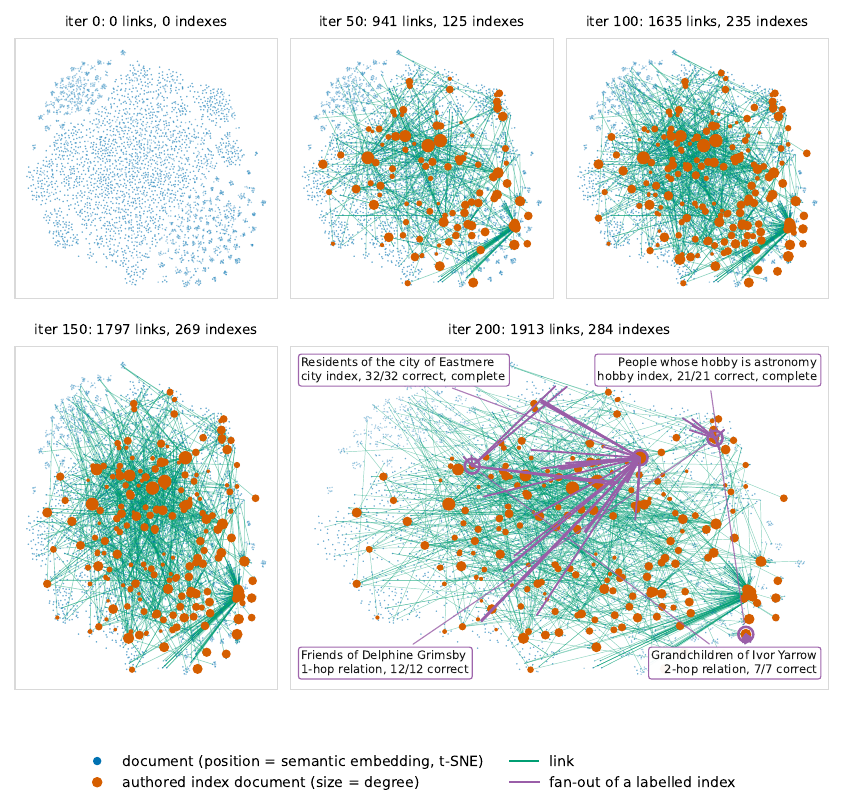}
\caption{The document network the agent built. \emph{Top:} the store at
three points in training; every document is drawn, and position is a t-SNE
projection of its embedding under the same function the store indexes with,
so documents drawn near each other are documents the reader's
\texttt{search} retrieves together (t-SNE preserves neighbourhoods, not
distances, so only local proximity should be read). One layout, computed
once on the final store and keyed by document, is shared by all panels, so a
document holds its position throughout and only the structure drawn over it
changes; the clusters visible at iteration~0 are the universe's natural
topic groups. Blue: source documents. Orange: authored index documents,
sized by degree. Green: links. The final panel (bottom right, at two thirds width) carries four
index documents labelled, together with the documents each one reaches
(purple). The four were selected for correctness, not size: every link each
of them carries points at a document that genuinely belongs under its key
(Sec.~\ref{sec:res-e4}), and together they span what the agent produced ---
an attribute index over a city, an attribute index over a hobby, a one-hop
relation, and a two-hop relation. Structure grows from nothing into a
navigation layer over the semantic space: hub indexes fan out into topic
clusters, and every link is index-mediated --- the agent never wires two
source documents to each other directly.}
\label{fig:network}
\end{figure*}

\begin{figure}[!t]
\centering
\includegraphics[width=0.92\columnwidth]{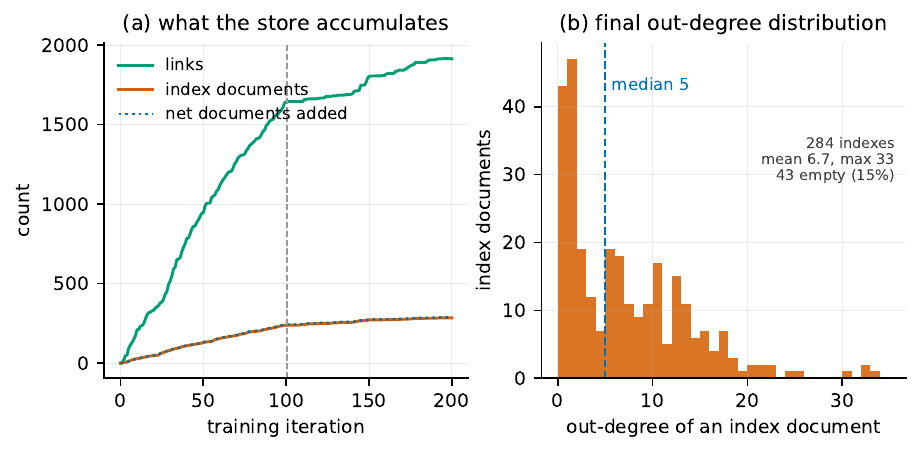}
\caption{(a)~What the store accumulates, reconstructed by trace replay.
\emph{Index documents} = agent-authored documents still alive; \emph{net
documents added} = store size minus the initial 5{,}864. The two curves
coincide: the agent adds access structure and essentially never rewrites
source documents, which is the parsimony objective realized (growth
$+4.9\%$ over 200 iterations). Link growth flattens after the epoch
boundary (dashed), the store saturating on the keys the training questions
name. (b)~Out-degree of every index document in the final store. The
distribution is long-tailed --- median~5, mean~6.7, max~33 --- with the
right tail carrying the attribute indexes of Table~\ref{tab:idxquality} and
a spike at zero: $43$ indexes ($15\%$) were opened and never filled.}
\label{fig:trajectory}
\end{figure}

\subsection{E4: What the Agent Built}\label{sec:res-e4}
\textbf{Shape and cost of curation.} Fig.~\ref{fig:network} maps the store's
evolution across its semantic space: structure grows from nothing into a
navigation layer, hub indexes fanning out into topic clusters. Trace replay
puts numbers on it (Fig.~\ref{fig:trajectory}a): $+287$ documents
($+4.9\%$) and $1{,}913$ links, nearly all laid down in epoch~1. The
backward pass is dominated by two actions --- $307$ \texttt{add} and $284$
\texttt{link\_many} calls across the run, against $3$ edits, $12$ deletes
and $9$ unlinks --- so the agent builds and almost never revises. The
deduplication judge fired on $45$ candidates and merged $8$; document-level
parsimony was already holding at this scale.

\textbf{Are they indexes?} Calling every authored document an index would
beg the question, so we classify the 287 by three checkable properties: at
least one link; a sentence that does not enumerate its own members (the
shape the curation skill forbids, tested by resolving the key's true member
set from the universe and looking for those names in the text); and a
resolvable key. On that test $242$ ($84.3\%$) are genuine indexes, $43$
($15.0\%$) are empty stubs like \emph{``Delphine Thistlewood''}, and $2$ are
materialized answers. That $84\%$ is itself a result: in an earlier version
of this environment the same protocol produced the opposite shape, $80\%$ of
authored documents stating the verified answer as a sentence, which serves
the one question that produced it. The difference is that the curation skill
now names the mechanism (``an index points; it does not list'') rather than
only the goal.

\begin{table}[!t]
\caption{Index construction quality, by the kind of key the index is built
on, with one example of each.}
\label{tab:idxquality}
\centering
\scriptsize
\renewcommand{\arraystretch}{1.1}
\setlength{\tabcolsep}{4pt}
\begin{tabular*}{\columnwidth}{@{\extracolsep{\fill}}lrrrr@{}}
\toprule
Key the index is built on & $n$ & deg. & prec. & recall \\
\midrule
attribute: city & 8 & 26.4 & 98\% & 95\% \\
\multicolumn{5}{@{}l}{\quad\emph{``Residents of Dorringham''}} \\[1pt]
attribute: hobby / job & 54 & 12.7 & 98\% & 96\% \\
\multicolumn{5}{@{}l}{\quad\emph{``People whose hobby is astronomy''}} \\[1pt]
single-entity hub & 91 & 8.0 & 95\% & 90\% \\
\multicolumn{5}{@{}l}{\quad\emph{``Bennett Coldwater''}} \\[1pt]
relation, two-hop & 10 & 3.9 & 87\% & 55\% \\
\multicolumn{5}{@{}l}{\quad\emph{``Grandchildren of Ivor Yarrow''}} \\[1pt]
relation, one-hop & 54 & 3.4 & 78\% & 73\% \\
\multicolumn{5}{@{}l}{\quad\emph{``Friends of Delphine Grimsby''}} \\
\midrule
all scored indexes & 220 & 8.5 & 94\% & 91\% \\
\bottomrule
\end{tabular*}
\end{table}

\textbf{Accurate, but narrow.} Table~\ref{tab:idxquality} scores every index
whose key resolves. A link is correct if it points at a document about a
genuine member of the key, resolved against the universe rather than by
string match --- the grandchildren of a person are recorded as ``\emph{X} is
a child of \emph{Y}'' and never name the grandparent, so a string-match test
scores a correct two-hop index at zero, and an earlier version of this
analysis reported exactly that artifact. Resolved properly, precision is
$94\%$ and member recall $91\%$ over the 220 whose key resolves. Quality tracks
the arity of the key, not its depth: attribute indexes over a city or hobby
are near-perfect ($98\%$, $95$--$96\%$ recall) and carry the most links,
while relational indexes are smaller and noisier and two-hop relations are
\emph{not} worse than one-hop ($87\%$ vs.\ $78\%$). Depth is not what the
curator struggles with; breadth is. Structurally the layer is flat:
$1{,}896$ of $1{,}913$ links attach an index, $100$ join two indexes and
$17$ join two source documents, so entry points are built readily and levels
almost never --- and \textbf{$1{,}621$ of $5{,}864$ source documents
($27.6\%$) are reachable from an index in one read}. That is the binding
constraint of Sec.~\ref{sec:res-grad}, restated as a property of the store.

\textbf{Break-even, and what would settle the mechanism.} On the trained
group the store saves $5{,}466$ answering tokens per question ($13.7$k
against $8.2$k), so the 18.8M-token run repays itself after $N^{*} \approx
3{,}400$ questions (Eq.~\ref{eq:breakeven}) --- a figure that prices
re-asking within the covered population, not generalization. We had intended
to attribute the step savings to hits on built structure by splitting exam
questions on whether the trajectory touched an index, and report that this
split does not identify: touching is \emph{downstream} of searching, so a
longer trajectory is more likely to encounter an index and the touched group
is selected for difficulty by construction. The gradient probe is the
intervention that does identify, because it varies coverage of the
question's key \emph{before} the reader starts.

\section{Discussion and Limitations}\label{sec:disc}
\textbf{Coverage, not construction, is the limit.} The indexes the agent
builds are accurate ($94\%$ precision, $91\%$ member recall); what it does
not build is \emph{enough} of them, and the quarter of the corpus it covers
is the quarter the training questions named. Two mechanisms are implicated,
both actionable: the backward budget cannot populate a thirty-member index
in one iteration and nothing asks the agent to \emph{return} to one, so $43$
were opened and abandoned; and more fundamentally the agent indexes the key
a question names rather than the class it belongs to --- ``Friends of
Delphine Grimsby'' when asked about her, never ``friends, for everyone''. A
protocol scoring an index by completeness over a class, and rewarding
extension over creation, is the obvious next experiment. Curation has a
boundary in the other direction too: GraphRAG-style summaries added to an
already-searchable store \emph{lose} accuracy, so the value is in knowing
when not to edit --- something downstream-only evaluation cannot see, since
a store can be slowly ruined while individual answers still look fine.

\textbf{Deployment and the online variant.} The abstraction targets agent
fleets maintaining shared repositories, where our failure ledger maps onto
real incidents: near-duplicates burying each other is the lost-update
problem, authored content without provenance is the hallucinated fix, a
deleted last instance is knowledge loss during refactoring. We separate
training from a frozen exam because that is what makes the generalization
question answerable, but nothing in the method requires it: the online
variant of Sec.~\ref{sec:res-perlink} needs a different \emph{measurement},
not a different curator, since with the store moving underneath held-out
accuracy stops being well defined and the honest alternative is prequential.

\textbf{Threats to validity.} Template-rendered language is simpler than
natural prose and may flatter lexical matching; the PhantomWiki arm, whose
generator we do not control, is the partial answer. Reader and curator share
a model family (\texttt{gpt-5-mini-2025-08-07}), so structure tuned by one
may suit the other's habits --- though a stronger reader is the harder test
for us, not the easier one, since the better it is at recovering a set by
searching the less an index adds. The four-point gradient is \kbgym{}-only:
it needs a generator that can enumerate a template's instance pool. Single
seed; single agent by design; no support-set diagnostics on the external
arm; grading is token-F1 against short golds.

\end{document}